\documentclass{article}
\usepackage{gensymb}
\usepackage{subfig}
\usepackage{multirow}
\usepackage{spconf,amsmath,graphicx}
\usepackage{amssymb}

\title{Contextual Pyramid Attention network for building segmentation \\ in aerial imagery}
\name{Clint Sebastian, Raffaele Imbriaco, Egor Bondarev, Peter H.N. de With}
\address{Eindhoven University of Technology, Eindhoven, The Netherlands}
\begin{document}
%
\maketitle
\begin{abstract}
Building extraction from aerial images has several applications in problems such as urban planning, change detection, and disaster management. With the increasing availability of data, Convolutional Neural Networks (CNNs)  for semantic segmentation of remote sensing imagery has improved significantly in recent years. However, convolutions operate in local neighborhoods and fail to capture non-local features that are essential in semantic understanding of aerial images. In this work, we propose to improve building segmentation of different sizes by capturing long-range dependencies using contextual pyramid attention (CPA). The pathways process the input at multiple scales efficiently and combine them in a weighted manner, similar to an ensemble model. The proposed method obtains state-of-the-art performance on the Inria Aerial Image Labelling Dataset with minimal computation costs. Our method improves 1.8 points over current state-of-the-art methods and 12.6 points higher than existing baselines on the Intersection over Union (IoU) metric without any post-processing. Code and models will be made publicly available.
\end{abstract}
\vspace{-5mm}
\section{Introduction}
\label{intro}
The developments in the systematic collection and organization of remote sensing imagery have resulted in several high-resolution aerial imagery datasets. Information from aerial imagery plays a key role in urban planning, disaster aversion, and change detection. Building detection is a crucial aspect for the aforementioned applications. Depending on the geographical region and conditions, building structures have different shapes and sizes. This challenge is particularly addressed by Maggiori \textit{et al.} \cite{maggiori2017can}. They created a dataset of labeled aerial imagery from different locations for this problem, such that a model trained from a variety of sources generalizes to the task of segmentation. Semantic segmentation in aerial imagery is challenging due to variable lighting conditions, shapes/sizes, and large intraclass variations. In this research, we address the problem of improving the building segmentation by utilizing attentive multi-scale pathways. Each of the paths exploits non-local neighborhoods that account for buildings of varying sizes. This allows our network to learn long-range dependencies at various scales with minimal computation costs. In addition to attentive multi-scale pathways, we also incorporate a channel-wise attention module to model interdependencies across channels.
In summary, our contributions are as follows. 

\begin{figure}[t]
    \centering
    \includegraphics[width=1.0\linewidth]{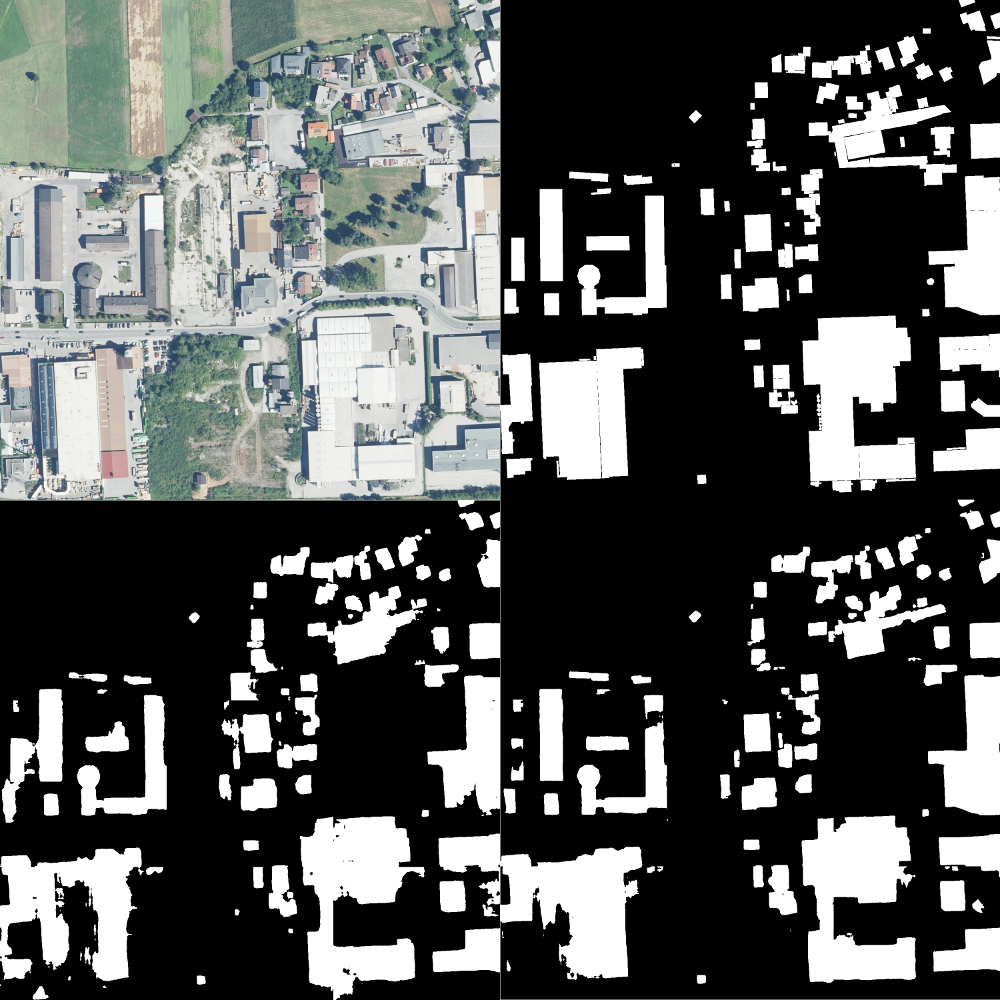}
    \caption{Comparison of building segmentation on Inria Aerial Image Labeling Dataset. RGB, GT and outputs (bottom row) ResNet101-FPN without (left) and with our module (right).}
    \label{fig:intro_picture}
\end{figure} 

\begin{itemize}
    \item We introduce a self-attention based contextual pyramid attention (CPA) module that accounts for various building sizes to segment buildings in aerial images. The proposed module outperforms current state-of-the-art methods by about 2\% on the IoU metric and 12.6\% over FCN baselines. 
    \item Through experiments, we also show that our base model offers competitive performance to current state-of-the-art methods, while having much lower inference costs. We also provide ablation studies on the impact of our proposed module and other comparisons. 

\end{itemize}

\section{RELATED WORK}
\label{related_work}
 Convolutional Neural Networks (CNNs) have improved the performance of semantic segmentation significantly in recent years. Encoder-Decoder architectures are a popular choice for segmentation. Using fully convolutional architectures for segmentation is proposed in \cite{long2015fully}. Architectures such as FCN, U-Net, DenseNet, etc., are often applied to obtain higher quality segmentations \cite{long2015fully, huang2016densely, jegou2017one}. Nowadays, most segmentation architecture designs are based on using a pre-trained backbone such as VGGNet, ResNet, and a complex decoder such as Pyramid Scene Parsing (PSP) or Atrous Spatial Pyramid Pooling (ASPP)\cite{simonyan2014very, he2015deep, xie2017aggregated, hu2018squeeze, zhao2017pyramid, chen2018encoder}.

Due to the unique nature of remote sensing imagery, several custom architectures and loss functions have been developed for aerial image building segmentation \cite{bischke2019multi, sebastian2020adversarial}. In \cite{bischke2019multi}, a multi-task learning approach is introduced, where a distance-transform loss function is applied in conjunction with the cross-entropy loss. Similar to our work, \cite{marcu2016dual} proposes to combine local and global features to improve semantic segmentation of buildings in aerial imagery. However, the combination of two VGGNets makes the overall inference computationally expensive. Our proposed approach circumvents this by splitting only the last convolutional block to operate at multiple scales. In \cite{Marcu2018AMM}, a joint multi-stage multi-task approach is used, where the first stage trains a segmentation network, and the second stage trains for geo-localization using a multi-task loss function. Apart from these, post-processing techniques such as Conditional Random Fields (CRFs) and test-time augmentations are applied to improve segmentation performance. In \cite{mou2018rifcn}, a recurrent network is applied in a fully convolutional network that exploits a decoder network fusing features from the encoder layer in a similar fashion to feature pyramid networks (FPN). An FPN utilizes a pyramidal hierarchy of features extracted from an encoder that is later combined with the decoder via lateral connections \cite{lin2017feature}. However, as FPNs utilize only a few of the earlier layers of the encoder for lateral connections, the features might not be rich as extracting from deeper layers.

A common practice to retain feature resolution is using dilated convolutions. Due to the larger feature resolution and the repeated object patterns in aerial imagery, capturing long-range relations is beneficial. However, a single-sized feature map may not be able to capture all objects of variable sizes. For example, spatial relationships across large buildings are easier to capture when feature maps have a low resolution. When buildings are smaller in size, larger feature resolution is suited to capture more fine-grained details. Using a decoder such as FPN is beneficial for high-resolution features, however, as they are not obtained from deeper layers, they lack rich semantics. The proposed approach in this research is suited for these properties that are typically exhibited in aerial imagery. We particularly address the problem of improving semantic segmentation of buildings at various sizes.

\section{METHOD}

\begin{figure*}
    \centering
    \includegraphics[width=\linewidth]{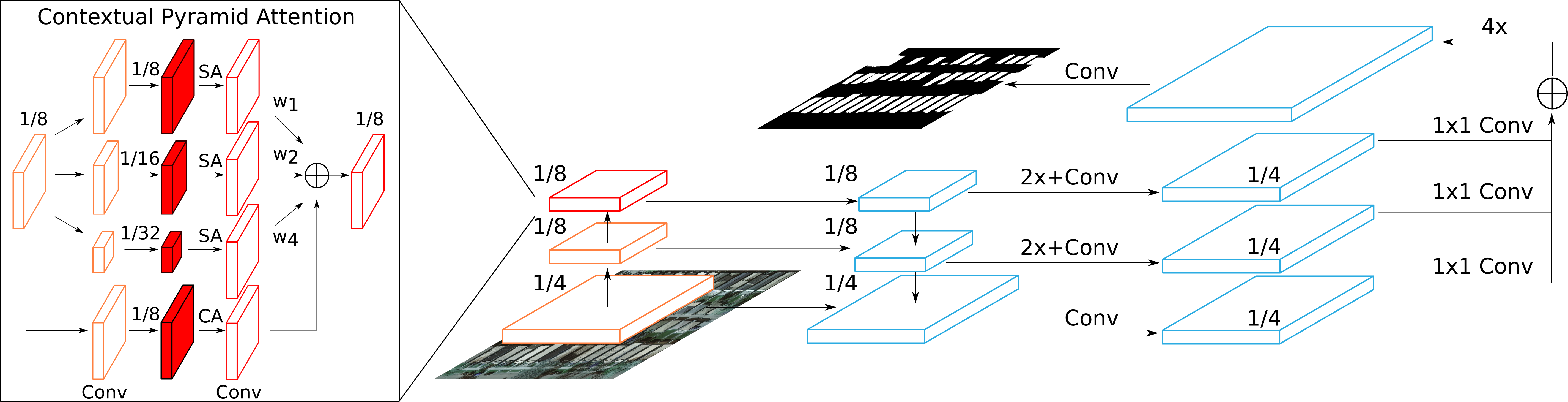}
    \caption{Overview of the proposed network. The network uses a ResNet backbone for the encoder and Feature Pyramid Network decoder. The proposed context pyramid attention (CPA) is applied after Block 4 of ResNet. CPA consists of contextual and channel-wise attention modules, where the contextual attention operates at multiple scales to produce a weighted output that is fed to the FPN decoder (CPA module at the left). }
    \label{fig:architecture}
\end{figure*}
\label{method}
The architecture is composed of an encoder and decoder, similar to other segmentation networks. An overview of the proposed method is shown in Fig.~\ref{fig:architecture}. Each component of our model is described in the following subsections.
\vspace{-4mm}
\subsection{Architecture}
\textbf{Encoder.} We consider three backbones for training, ResNet18, ResNet101 and Squeeze and Excitation ResNeXt101 (SEResNeXt101). ResNet18 serves as the light architecture that may be suited for fast inference, whereas SEResNeXt101 offers higher performance with additional computation cost. SEResNeXt101 incorporates an additional Squeeze and Excitation module along with aggregated residual connections.
\newline
\newline
\textbf{Decoder.} The decoder is a Feature Pyramid Network (FPN), similar to Semantic FPN in \cite{kirillov2019panoptic}. FPN combines features at different spatial resolutions through lateral connections to a top-down decoder. The lateral connections are established by combining the bottom-up outputs at each corresponding level of the top-down decoder. Each of the combined top-down decoder features is upsampled to $1/4^{th}$ of the input resolution and is combined via summation to produce rich features that are transformed into a segmentation mask.
\vspace{-3mm}
\subsection{Contextual Pyramid Attention block}
The proposed Contextual Pyramid Attention block is composed of two parts, a contextual attention module that captures long-range spatial dependencies and a channel-wise attention module. Both proposed attention modules are based on self-attention~\cite{vaswani2017attention, nam2017dual}. 
\newline
\newline
\textbf{Contextual attention.} Buildings in aerial imagery appear in various sizes, and structural features are often redundant in a given region (e.g. repeated houses structure in a neighborhood). This non-local information is not immediately processed by convolutions as they operate in a limited region defined by the kernel size. Self-attention, on the other hand, can bring in the understanding of long-range dependencies capturing the relation between repeated structures. It is a standard practice in semantic segmentation to use dilated convolutions in deeper layers, to increase the feature resolution. This is beneficial for smaller buildings and furthermore, self-attention blocks can capture fine-grained relations across the image. However, the contextual information of buildings of different sizes becomes challenging for two reasons. First, the dilated convolutions retain the feature resolution and self-attention may then retain spatial relations that are too fine-grained for larger buildings, which may not provide sufficient spatial context. Second, if the dilated convolutions are not applied to retain feature resolution, this may fail to capture relations between smaller buildings. Our motivation using multi-scale pathways stems from these two reasons.
\newline
\textbf{Method.} Given a feature $\mathbf{F} \in \mathbb{R}^{C \times H \times W}$, we apply convolutions to generate key, query and value features $\mathbf{K}, \mathbf{Q}, \mathbf{V} \in \mathbb{R}^{C \times H \times W}$. The tensors $\mathbf{K}, \mathbf{Q}$ and $\mathbf{V}$ are reshaped into $N \times C$, where $N$~=~$H~\times~W$. The self-attention (SA) operation is defined as
\begin{equation}
    \mathbf{A}_{s} = \text{SA}(\mathbf{F})= \gamma_{s} \cdot (\text{Softmax}(\mathbf{K}\mathbf{Q}^{T}))\mathbf{V},
\end{equation}
where $\mathbf{A}_{s} \in \mathbb{R}^{C \times N}$ is then reshaped into $C \times H \times W$ and $\gamma_s$ is a learned parameter. The output of the self-attention is the sum of $\mathbf{A}_{s}$ and $\mathbf{F}$. The resolution of $\mathbf{F}$ is retained by dilated convolutions ($1/8^{th}$ of the input image resolution). To capture contextual long-range dependencies, we operate the deep features $\mathbf{F}$ at various scales $s$. Contextual attention is
\begin{equation}
    \mathbf{C}_{s} = \sum_{s \in (1, 2, 4)} w_{s} \cdot \text{Conv}((\mathbf{F}_{1/s} + \text{SA}(\text{Conv}(\mathbf{F}_{1/s})))_{s}),
\end{equation}
where Conv is the convolution operation, followed by ReLU activation and Batch Normalization \cite{nair2010rectified, ioffe2015batch}. The features are downsampled by a factor of $1/s$, followed by Conv and SA operations. The resultant features are reshaped and upsampled to the original size, followed by another Conv operation. Each of these paths is weighted by a parameter $w_s$. The sum of the pathways results in the contextual attention output $\mathbf{C}_s$.  
\newline
\newline
\textbf{Channel-wise attention.} Contextual attention employs both spatial axes to establish interdependencies to model spatial information. However, higher-level class or object information is prevalent across channels, and hence, we apply channel-wise attention to model relations across them. The last feature block of ResNet is large with 2,048 channels and is computationally expensive to perform self-attention based operations. Therefore, the features are compressed to $1/4$ of the total number of channels by 1 $\times$ 1 convolutions to obtain the features $\mathbf{F}_{1/c}$, where $c$ is the compression factor across channels (note the difference with scale factor $s$). The key, query and value are all the same in this case ($\mathbf{F}_{1/c}$). However, to apply self-attention, the features are reshaped into $C \times N$, where $N$ = $H \times W$. The channel-wise attention is now
\begin{equation}
    \mathbf{A}_{cw} = \gamma_c \cdot( \text{Softmax}(\mathbf{F}_{1/c}\mathbf{F}_{1/c}^T))\mathbf{F}_{1/c}.
\end{equation}
The final output is the sum of $\mathbf{F}_{1/c}$ and $\mathbf{A}_{cw}$. Note that the affinity matrices (inputs to the softmax) generated through both attention mechanisms have different shapes. The contextual attention affinity matrix has a size of $N \times N$, whereas the channel-wise affinity matrix has a size of $C \times C$. The final contextual pyramid attention block is the sum of both contextual and channel-wise attention.
\section{EXPERIMENTS}
\label{experiments}

\begin{table*}%

\begin{minipage}{0.72\linewidth}
  \footnotesize
    \begin{tabular}[t]{l|c|c|c|c|c|c|c} 
    \hline
    Method &  & Austin & Chicago & Kitsap Co. & West Tyrol & Vienna & Overall \\ 
    \hline
    \hline
    FCN + MLP \cite{maggiori2017can} & IoU & 61.20 & 61.30 & 51.50 & 57.95 & 72.13 & 64.67 \\
     & Acc. & 94.20 & 90.43 & 98.92 & 96.66 & 91.87 & 94.42 \\ 
    \hline
    SegNet MT-Loss  \cite{bischke2019multi} & IoU & 76.76 & 67.06 & \textbf{73.30} & 66.91 & 76.68 & 73.00 \\
     & Acc. & 93.21 & \textbf{99.25} & 97.84 & 91.71 & 96.61 & 95.73 \\ 
    \hline
    MSMT-Stage-1  \cite{marcu2018multi} & IoU & 75.39 & 67.93 & 66.35 & 74.07 & 77.12 & 73.31 \\
     & Acc. & 95.99 & 92.02 & 99.24 & 97.78 & 92.49 & 96.06 \\ 
    \hline
    2-levels U-Net  + & IoU & 77.29 & 68.52 & 72.84 & 75.38 & 78.72 & 74.55 \\
    aug. \cite{khalel2018automatic} & Acc. & 96.69 & 92.40 & 99.25 & 98.11 & 93.79 & 96.05 \\ 
    \hline
    ICT-Net \cite{chatterjee2019semantic} & IoU & - & - & - & - & - & 75.50 \\
     & Acc. & - & - & - & - & - & 96.05 \\
    \hline
    \hline
    ResNet18-FPN  & IoU & 77.89 & 69.73 & 65.65 & 76.95 & 79.49 & 75.51 \\
    -CPA (ours) & Acc. & 96.77 & 92.70 & 99.25 & 98.13 & 94.09 & 96.19\\
    \hline
    ResNet101-FPN & IoU & 78.93 & \textbf{71.57} & 68.06 & 78.29 & 81.03 & 77.03\\
    -CPA (ours) & Acc. & 96.98 & 93.20 & 99.27 & 98.29 & 94.61 & 96.47\\
    \hline
    SEResNeXt101-FPN & IoU & \textbf{80.15} & 69.54 & 70.36 & \textbf{80.83} & \textbf{81.43} & \textbf{77.29} \\
    -CPA (ours) & Acc. & \textbf{97.18}  & 92.78 & \textbf{99.32} & \textbf{98.46} & \textbf{94.67} &  \textbf{96.48}\\
    \hline
    \end{tabular}
  \end{minipage}\hfill
  \begin{minipage}{0.28\linewidth}
        \footnotesize
        \begin{tabular}[t]{l|c|c}
        \hline
        \textbf{Model} & \textbf{Accuracy} & \textbf{IoU} \\ \hline \hline
        ResNet18-FPN & 95.9 & 73.97 \\ \hline
        -Self-attention & 96.0 & 74.64 \\ \hline
        -CPA (ours)  & 96.2 & 75.51 \\ \hline 
        \end{tabular}%

        \qquad

        \begin{tabular}[t]{l|c|c}   
        \hline
        \textbf{Model} & \textbf{Time/tile} & \textbf{IoU} \\ \hline \hline
        ResNet18 & 4.67 s & 75.51 \\ \hline
        ResNet101 & 12.16 s & 77.03 \\ \hline
        SE-ResNeXt101 & 14.07 s & 77.29 \\ \hline
        \end{tabular}%
        
        \qquad
        \newline
        \newline
        \includegraphics[width=1.0\linewidth]{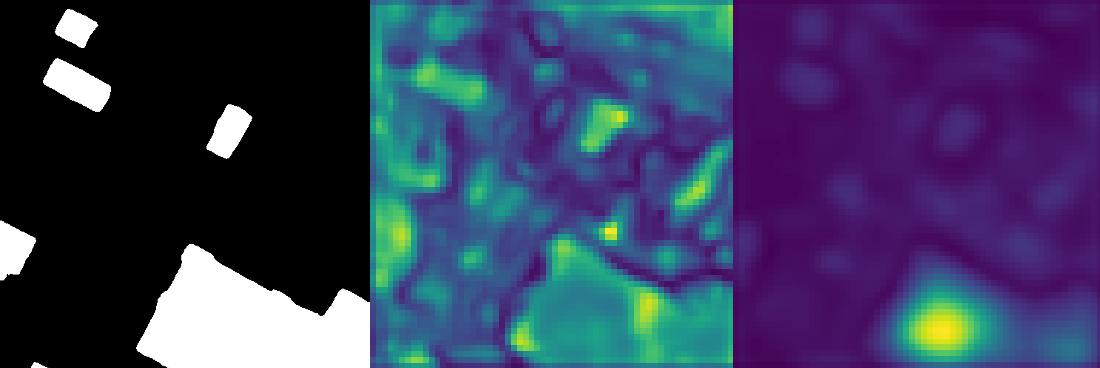}
  \end{minipage}
\captionsetup[table]{labelformat=bold}
\caption{(Left) Performance comparison of CPA with other state-of-the-art methods. (Top-right) Comparison of our method with/without the CPA module. (Mid-right) Inference time (in sec.) per tile of 5000 $\times$ 5000 pixels with different backbones with CPA module. (Bottom-right) Prediction, high- and low-resolution attention map. The high-resolution learns semantics related to small buildings and fine-grained details of large buildings whereas low resolution learns semantics related to large buildings.}%
  \label{tbl:table}
  \vspace{-2mm}
\end{table*}

\subsection{Dataset and Evaluation}
\textbf{Dataset.} The Inria Aerial Image Labeling Dataset consists of 360 RGB ortho-rectified aerial images at a spatial resolution of 0.3 meters. Each of the ortho-rectified images has a resolution of 5000 $\times$ 5000 pixels, covering a region of 1500 square meters per image. The dataset comprises of 10 different cities, out of which 5 cities are available for training. The images are taken at different urban conditions from various cities in Europe and America. The groundtruth is available only for the training set, where information is available for two classes: buildings and non-buildings. For fair comparisons, we follow the same evaluation protocol for testing as in \cite{bischke2019multi, marcu2018multi, khalel2018automatic}. Images 1--5 of each location are used for validation and Images 6--36 for training.
\newline
\newline
\textbf{Evaluation.} For evaluation, we use both Intersection over Union (IoU) and accuracy. IoU is measured across the building class. IoU is the ratio of true positive pixels to the pixels that are labeled as positive in the ground-truth or predictions. We also utilize accuracy, which is the ratio of correctly classified pixels to the total. Each dataset is evaluated separately and jointly, for both of these metrics.
\newline
\textbf{Implementation details.} Each of the networks are initialized with weights pretrained on ImageNet~\cite{deng2009imagenet}. The models are trained using the Adam optimizer with a learning rate of 0.00001~\cite{kingma2014adam}. The objective function is cross-entropy and is trained for 35 epochs. The input images have a size of 500 $\times$ 500 pixels and are randomly rotated by 90\degree or flipped horizontally and vertically while training. Images are also augmented with minor changes in brightness and contrast. The image augmentations are applied using Albumentations~\cite{2018arXiv180906839B}, and the models are implemented in PyTorch. Each of the contextual and self-attention weights are intialized to 1 and 0.05, respectively. 
\vspace{-3mm}
\subsection{Comparison with state-of-the-art}
With the proposed module, we observe an improvement of 3-5\% IoU across all the datasets except Kitsap Co.. Our method with a simple backbone such as ResNet18 offers competitive performance to current state-of-the-art results. Using a deeper backbone such as ResNet101 or SE-ResNeXt101 further improves performance. Compared to the previous state-of-the-art ICT-Net that uses U-Net with Dense Blocks and SE Blocks~\cite{hu2018squeeze}, we observe a gain of 1.8 IoU points. The improvement is higher than previous increases in recent years. Our method does not rely on post-processing, or multi-task learning methods, however, applying this may further improve performance. A visual comparison of our results is in Fig. \ref{fig:intro_picture}.
We note that on Kitsap Co., which contains the fewest buildings, our performance decreases slightly. We observe very high accuracy (99.32\%), whereas IoU is lower, indicating skewness to classifying background better. On visual inspection, we have noticed errors in the ground truth where the background is annotated as buildings. The incorrect ground truth, when combined with the sparsely covered building regions, penalizes the IoU metric, thereby lowering performance.

\subsection{Ablation studies}
To study the impact of attention, an ablation study is conducted without any attention, with self-attention and the proposed CPA. This is shown in Table~\ref{tbl:table} (top-right). The IoU improves by 0.87 and 1.54 points with the addition of self-attention and CPA over the ResNet-FPN. Furthermore, we conduct experiments to study the impact of different models on computation costs (Table~\ref{tbl:table} mid-right). For a tile of 5000 $\times$ 5000 pixels, ResNet18-FPN-CPA takes only 4.67 seconds on a GTX 1080Ti GPU, whereas ResNet101 and SE-ResNeXt101 take 12.16 and 14.07 seconds, respectively. In \cite{khalel2018automatic}, a tile takes 160 seconds for processing, however, they use a K80 GPU, which is typically two times slower than our setup. Even so, compared to \cite{khalel2018automatic}, our largest model (SE-ResNeXt101) is five times faster for inference, while offering higher performance.
\vspace{-2mm}
\section{CONCLUSIONS}
\label{conclusions}
We have presented a novel and effective method to improve building segmentation in aerial imagery. It consists of the contextual pyramid and channel-wise attention blocks to model long-range dependencies across spatial contexts to account for buildings of different sizes. The contextual pyramid attention combines contextual information at multiple scales in an efficient manner with minimal computation overhead, whereas channel-wise attention captures interdependencies across channels. The proposed method achieves state-of-the-art performance on the Inria Aerial Image Labelling dataset by 1.8 and 12.6 IoU points over previous state of the art and baseline. Furthermore, we perform ablation experiments to study the impact of the CPA module and provide comparisons for computation costs. Our model offers high performance without any post-processing with low inference times.

\small
\bibliographystyle{IEEEbib}
\bibliography{strings,refs}

\end{document}